\documentclass{article}

\usepackage{verbatim}
\usepackage{amssymb}
\usepackage{amsbsy}
\usepackage{amscd}
\usepackage{amsmath}
\usepackage{amsthm}
\usepackage[mathscr]{eucal}

\newcommand{\R}{\mathbb R}

\newtheorem{theorem}{Theorem}
\newtheorem{defn}{Definition}\numberwithin{defn}{section}
\newtheorem{lemma}{Lemma}

\title{A Limit Theorem in Singular Regression Problem}

\author{Sumio Watanabe\\
Precision and Intelligence Laboratory\\
Tokyo Institute of Technology\\
4259 Nagatsuta, Midoriku, Yokohama, 226-8503AJapan\\
e-mail:swatanab@pi.titech.ac.jp
}

\begin{document}

\maketitle

\begin{abstract}
In statistical problems, a set of parameterized 
probability distributions is used to estimate 
the true probability distribution. If Fisher 
information matrix at the true distribution is singular, 
then it has been left unknown what we can estimate about the 
true distribution from random samples. 
In this paper, we study a singular regression problem and 
prove a limit theorem which shows the relation between 
the singular regression problem and two birational invariants, 
a real log canonical threshold and a singular fluctuation. 
The obtained theorem has an important application to statistics, 
because it enables us to estimate the generalization error from the
training error 
without any knowledge of the true probability distribution. 
\end{abstract}

\section{Introduction}

Let $M$ and $N$ be natural numbers, and ${\R}^{M}$ and ${\R}^{N}$ be 
$M$ and $N$ dimensional real Euclidean spaces respectively. 
Assume that $(\Omega,{\mathcal B},P)$ is a probability space and that 
$(X,Y)$ is an ${\R}^{M}\times{\R}^{N}$-valued
random variable which is subject to
a simultaneous probability density function, 
\[
q(x,y)=\frac{q(x)}{(2\pi\sigma^{2})^{N/2}}
\exp\Bigl(-\frac{|y-r_{0}(x)|^{2}}{2\sigma^{2}}\Bigr),
\]
where $q(x)$ is a probability density function on ${\R}^{M}$,
$\sigma>0$ is a constant, $r_{0}(x)$ is a measurable function from ${\R}^{M}$ to
${\R}^{N}$, and $|\cdot|$ is the Euclidean norm of ${\R}^{N}$. 
The function $r_{0}(x)$ is called a regression function of $q(x,y)$. 
Assume that $\{(X_{i},Y_{i});i=1,2,...,n\}$ is a set of random variables
which are independently subject to the same probability distribution
as $(X,Y)$. Let $W$ be a subset of ${\R}^{d}$. 
Let $r(x,w)$ be a function from ${\R}^{M}\times W$ to
${\R}^{N}$. The square error $H(w)$ is a real function on $W$, 
\[
H(w)=\frac{1}{2}\sum_{i=1}^{n}|Y_{i}-r(X_{i},w)|^{2}.
\]
An expectation operator $E_{w}[\;\;\;]$ on $W$ is defined by
\begin{equation}\label{eq:Ew}
E_{w}[F(w)]=\frac{\displaystyle \int F(w)\exp(-\beta H(w))\varphi(w)dw}
{\displaystyle \int \exp(-\beta H(w))\varphi(w)dw},
\end{equation}
where $F(w)$ is a measurable function, 
$\varphi(w)$ is a probability density function on $W$, and 
$\beta>0$ is a constant called an inverse temperature. 
Note that $E_{w}[F(w)]$ is not a constant but a random variable
because $H(w)$ depends on random variables. 
Two random variables $G$ and $T$ are defined by
\begin{eqnarray*}
G&=&\frac{1}{2}E_{X}E_{Y}[|Y-E_{w}[r(X,w)]|^{2}],\\
T&=&\frac{1}{2n}\sum_{i=1}^{n}|Y_{i}-E_{w}[r(X_{i},w)]|^{2}.
\end{eqnarray*}
These random variables $G$ and $T$ are called the generalization 
and training errors respectively. 
Since $E_{X,Y}[|Y-r_{0}(X)|^{2}]=N\sigma^{2}$, it is expected 
on some natural conditions that 
both $E[G]$ and $E[T]$ converge to $S=N\sigma^{2}/2$ 
when $n$ tends to infinity 
if there exists $w_{0}\in W$ such that $r(x,w_{0})=r_{0}(x)$. 
In this paper, 
we ask how fast such convergences are, in 
other words, our study concerns with a limit theorem 
which shows the convergences 
$n(E[G]-S)$ and $n(E[T]-S)$, when $n\rightarrow\infty$.
If Fisher information matrix 
\[
I_{ij}(w)=\int \partial_{i}r(x,w)\cdot\partial_{j}r(x,w)q(x)dx,
\]
where $\partial_{i}=(\partial/\partial w_{i})$,
is positive definite for arbitrary $w\in W$, then this problem 
is well known as a regular regression problem. In fact, 
in a regular regression problem, convergences 
$
n(E[G]-S)\rightarrow d\sigma^{2}/2$ and 
$n(E[T]-S)\rightarrow  -d\sigma^{2}/2 $ hold.
However, if $I(w_{0})=\{I_{ij}(w_{0})\}$ is singular, that is to say,
if $\det I(w_{0})=0$, then the problem is called a singular regression problem
and convergences of $n(E[G]-S)$ and $n(E[T]-S)$ have been left unknown. 

In general it has been difficult to study a limit theorem 
for the case when Fisher information matrix is
singular. However, recently, we have shown that a limit theorem 
can be established based on resolution of singularities, and that 
there are mathematical relations between the limit theorem 
and two birational invariants in singular density estimation
\cite{watanabe2001, watanabe2006, watanabe2009}.
In this paper we prove a new limit theorem for the singular regression problem, which
enables us to estimate birational invariants from random samples. 
The limit theorem proved in this paper has an important application to 
statistics, because the expectation value of the 
generalization error $E[G]$ can be estimated from that of
the training error $E[T]$ without any knowledge of the true probability distribution. 
\vskip3mm\noindent
{\bf Example} Let $M=N=1$, $d=4$, $w=(a,b,c,d)$, 
and $W=\{w\in {\R}^{4};|w|\leq 1\}$. 
If the function $r(x,w)$ is defined by 
\[
r(x,w)=a\sin(bx)+c\sin(dx),
\]
and $r_{0}(x)=0$, then the set $\{w\in W;r(x,w)=r_{0}(x)\}$ is not one point, and
Fisher information matrix at $(a,b,c,d)=(0,0,0,0)$ 
is singular. A lot of functions used in statistics, information science,
brain informatics, 
and bio-informatics are singular, for example, artificial neural networks,
radial basis functions, and wavelet functions. 

\section{Main Results}

We prove the main theorems based on the following assumptions. 
\vskip3mm\noindent
{\bf Basic Assumptions}.\\
(A1) The set of parameters $W$ is defined by
\[
W=\{w\in {\R}^{d};\pi_{j}(w)\geq 0\;\;(j=1,2,...,k)\},
\]
where $\pi_{j}(w)$ is a real analytic function. 
It is assumed that $W$ is a compact set in ${\R}^{d}$ whose
open kernel is not the empty set. 
The probability density function $\varphi(w)$ on $W$ is given by 
\[
\varphi(w)=\varphi_{1}(w)\varphi_{2}(w),
\]
where $\varphi_{1}(w)\geq 0$ is a real analytic function
and $\varphi_{2}(w)>0$ is a function of class $C^{\infty}$. \\
(A2) Let $s\geq 8$ be the number that is equal to 4 times of some integer. 
There exists an open set $W^{*}\supset W$ such that
$r(x,w)-r_{0}(x)$ is an $L^{s}(q)$-valued analytic function on $W^{*}$, where
$L^{s}(q)$ is a Banach space defined by using its norm $|\;\;|_{s}$, 
\[
L^{s}(q)=\{f;|f|_{s}=\Bigl(\int |f(x)|^{s}q(x)dx\Bigl)^{1/s}<\infty\}. 
\]
(A3) There exists a parameter $w_{0}\in W$ such that $r(x,w_{0})=r_{0}(x)$. 
\vskip3mm\noindent
If these basic assumptions are satisfied, then 
\begin{equation}\label{eq:K(w)}
K(w)=\frac{1}{2}\int|r(x,w)-r_{0}(x)|^{2}q(x)dx
\end{equation}
is a real analytic function on $W^{*}$. A subset $W_{a}\subset W$ is defined by
\[
W_{a}=\{w\in W\;;\;K(w)\leq a\}.
\]
Note that $W_{0}$ is the set of all points that satisfy $K(w)=0$. 
In general, $W_{0}$ is not one point and it contains singularities. 
This paper gives a limit theorem for such a case. 
Proofs of lemmas and theorems in this section are given in section
\ref{section:proof}. 

\begin{lemma} \label{lemma:zeta}
Assume (A1), (A2), and (A3) with $s\geq 4$. Then
\[
\zeta(z)=\int_{W} K(w)^{z}\varphi(w)dw
\]
is a holomorphic function on $Re(z)>0$ which can be 
analytically continued to the unique meromorphic function 
on the entire complex plane whose poles are all 
real, negative, and rational numbers. 
\end{lemma}

\begin{lemma} \label{lemma:sf}
Assume (A1), (A2), and (A3) with $s\geq 8$. Then
there exists a constant $\nu=\nu(\beta)\geq 0$ such that 
\[
V=\sum_{i=1}^{n}\Bigl(E_{w}[\;|r(X_{i},w)|^{2}\;]-|\;E_{w}[r(X_{i},w)]\;|^{2}\Bigr)
\]
satisfies 
\begin{equation}\label{eq:E[V]}
\lim_{n\rightarrow\infty}E[V]=\frac{2\nu}{\beta}.
\end{equation}
\end{lemma}

Based on Lemma \ref{lemma:zeta} and \ref{lemma:sf}, we define two 
important values $\lambda,\nu>0$. 

\begin{defn}
Let the largest pole of $\zeta(z)$ be $(-\lambda)$ and
its order $m$. The constant $\lambda>0$ is called a real log canonical threshold. 
The constant $\nu=\nu(\beta)$ is referred to as a singular fluctuation. 
\end{defn}

The real log canonical threshold is an important invariant of an
analytic set $K(w)=0$. For its relation to algebraic geometry and algebraic analysis, 
see \cite{Atiyah,Bernstein,Gelfand,Mustata,Oaku1,Saito}. It is also important in
statistical learning theory, and it can be 
calculated by resolution of singularities \cite{watanabe2001,Aoyagi}. 
The singular fluctuation is an invariant of $K(w)=0$ 
which is found in statistical learning theory
\cite{WCCI2008,watanabe2009}, whose relation to singularity theory is still unknown. 
The followings are main theorems of this paper. 

\begin{theorem} \label{theorem:G-T}
Assume the basic assumptions (A1), (A2), and (A3) with $s\geq 8$. Let $S=N\sigma^{2}/2$.
Then 
\begin{eqnarray}
\displaystyle\label{eq:nE[G]}
\lim_{n\rightarrow\infty} n(E[G]-S)&=& \frac{\lambda-\nu}{\beta}+\nu\sigma^{2},\\
\displaystyle\label{eq:nE[T]}
\lim_{n\rightarrow\infty} n(E[T]-S)&= & \frac{\lambda-\nu}{\beta}-\nu\sigma^{2}.
\end{eqnarray}
\end{theorem}
This theorem shows that both the real log canonical
threshold $\lambda$ and singular fluctuation $\nu$ determine 
the singular regression problem.  

\begin{theorem} \label{theorem:eq-state}
Assume the basic assumptions (A1), (A2), and (A3) with $s\geq 12$. 
Then
\[
E[G]=E\Bigl[\Bigl(1+\frac{2\beta V}{nN}\Bigr)T\Bigr]+o_{n},
\]
where $o_{n}$ is a function of $n$ which satisfies $no_{n}\rightarrow 0$. 
\end{theorem}
By this theorem, 
$V$ and $T$ can be calculated from random samples without any direct knowledge
of the true regression function $r_{0}(x)$. Therefore, 
$E[G]$ can be estimated from random samples, resulting that we can find 
the optimal model or hyperparameter for the smallest generalization error. 
If the model is regular, then $\lambda=\nu=d/2$ for arbitrary $0<\beta\leq\infty$, 
resulting that Theorem \ref{theorem:eq-state}
coincides with AIC \cite{Akaike} of a regular statistical model. Therefore, 
Theorem \ref{theorem:eq-state} is a widely applicable information criterion, which 
we can apply to both regular and singular problems. We use
Theorem \ref{theorem:eq-state} without checking that the true distribution is 
a singularity or not. 

\section{Preparation of Proof}

We use notations, $S=N\sigma_{2}/2$ and 
\begin{eqnarray*}
S_{i}&=&Y_{i}-r_{0}(X_{i}),\\
f(x,w)&=&r(x,w)-r_{0}(x).
\end{eqnarray*}
Then $\{S_{i}\}$ are independent random variables which 
are subject to the normal distribution with average zero and
covariance matrix $\sigma^{2}I$ where $I$ is the $d\times d$ identity matrix. 
It is immediately derived that
\begin{eqnarray*}
E[T]&=&S-E\Bigl[\frac{1}{n}\sum_{i=1}^{n}S_{i}\cdot
E_{w}[f(X_{i},w)] \Bigr] \\
&& +E\Bigl[\frac{1}{2n}\sum_{i=1}^{n}|E_{w}[f(X_{i},w)]|^{2}\Bigr], \\
\\
E[G]&=&S+\frac{1}{2}E[E_{X}[|E_{w}[f(X,w)]|^{2}]], \\
E[V]&=&E\Bigl[\sum_{i=1}^{n}\{E_{w}[|f(X_{i},w)|^{2}]-|E_{w}[f(X_{i},w)]|^{2}\}\Bigr].
\end{eqnarray*}
The function $f(x,w)$ is an $L^{s}(q)$-valued
analytic function on $W^{*}$. In 
eq.(\ref{eq:Ew}), we can define $E_{w}[\;\;]$ by replacing $H(w)$ by $H_{0}(w)$, 
\[
H_{0}(w)=\frac{1}{2}\sum_{i=1}^{n}|f(X_{i},w)|^{2}-\sum_{i=1}^{n}S_{i}\cdot f(X_{i},w),
\]
which can be rewritten as
\[
H_{0}(w)= nK(w) - \sqrt{n}\;\eta_{n}(w),
\]
where $K(w)$ is given in eq.(\ref{eq:K(w)}), and 
\begin{eqnarray*}
\eta_{n}(w)&=& \eta_{n}^{(1)}(w)+\eta_{n}^{(2)}(w), \\
\eta_{n}^{(1)}(w)&=&
\frac{1}{\sqrt{n}}\sum_{i=1}^{n}S_{i}\cdot f(X_{i},w), \\
\eta_{n}^{(2)}(w)&=&\frac{1}{\sqrt{n}}\sum_{i=1}^{n}(K(w)-\frac{1}{2}|f(X_{i},w)|^{2}).
\end{eqnarray*}
We define a norm $\|\;\;\|$ of a function of $f$ on $W$ by
\[
\|f\|=\sup_{w\in W} |f(w)|.
\]
Since $W$ is a compact set of ${\R}^{d}$, the set
$B(W)$ that is a set of all
continuous and bounded function on $W$ 
is a Polish space, and both $\eta_{n}^{(1)}(w)$ and $\eta_{n}^{(2)}(w)$ 
are $B(W)$-valued random variables. Because $f(X,w)$ is an $L^{s}(q)$-valued
analytic function, $\{\eta_{n}^{(1)}\}$ and $\{\eta_{n}^{(2)}\}$ are
tight random processes, resulting that 
$\eta_{n}^{(1)}$ and $\eta_{n}^{(2)}$ weakly
converge to unique tight gaussian processes 
$\eta^{(1)}$ and $\eta^{(2)}$ respectively
which have the same covariance matrices as $\eta_{n}^{(1)}$ and $\eta_{n}^{(2)}$ 
respectively when $n\rightarrow\infty$ \cite{Wvan,watanabe2006,watanabe2009}. 

\begin{lemma} \label{lemma:eta-bdd}
Assume (A1), (A2), and (A3) with $s\geq 8$. Then 
\begin{eqnarray*}
E[\|\eta_{n}^{(1)}\|^{s}]&<&\infty,\\
E[\|\eta_{n}^{(2)}\|^{s/2}]&<&\infty,
\end{eqnarray*}
\end{lemma}
\begin{proof}
Since $f(x,w)$ is an $L^{s}(q)$-valued analytic function, it is represented by
the absolutely convergent power series $f(x,w)=\sum_{j} a_{j}(x)w^{j}$ which satisfies 
$|a_{j}(x)|\leq M(x)/r^{j}$ for some function $M(x)\in L^{s}(q)$ where 
$r=(r_{1},..,r_{d})$ is the associative convergence radii. 
By using this fact, the former inequality is proved \cite{watanabe2006,watanabe2009}. 
Also $K(w)-(1/2)f(x,w)^{2}$ is an $L^{s/2}(q)$-valued analytic function,
the latter inequality is proved. 
\end{proof}

\begin{lemma}\label{lemma:part} For arbitrary natural number $n$, 
\begin{eqnarray*}
E[E_{w}[\sqrt{n}\;\eta_{n}^{(1)}(w)]]&=&\sigma^{2}\beta E[V],\\
E[E_{w}[\sqrt{n}\;\eta_{n}^{(2)}(w)]]&=&
E[E_{w}[nK(w)-\frac{1}{2}\sum_{i=1}^{n}|f(X_{i},w)|^{2}]].
\end{eqnarray*}
\end{lemma}
\begin{proof}
The second equation is trivial. Let us prove the first equation.
Let the left hand side of the first equation be $A$. Since
$\{S_{i}\}$ are independently subject to the normal
distribution with covariance matrix $\sigma^{2}I$, 
\begin{eqnarray*}
A
&=& E\Bigl[\sum_{i=1}^{n}S_{i}\cdot E_{w}[f(X_{i},w)]\Bigr] \\
&=& \sigma^{2}E\Bigl[\sum_{i=1}^{n}\nabla_{S_{i}}\cdot E_{w}[f(X_{i},w)]\Bigr] \\
&=& \sigma^{2}E\Bigl[\sum_{i=1}^{n}\nabla_{S_{i}}\cdot 
\Bigl(
\frac{\int f(X_{i},w) \exp(-\beta H_{0}(w))\varphi(w)dw}
{\int \exp(-\beta H_{0}(w))\varphi(w)dw}\Bigr) \Bigr] \\
&=&\beta \sigma^{2}  E\Bigl[\sum_{i=1}^{n}
E_{w}[|f(X_{i},w)|^{2}]-|E_{w}[f(X_{i},w)]|^{2}\Bigr],
\end{eqnarray*}
which is equal to the right hand side of the first equation. 
\end{proof}

\begin{defn}
Let us define five random variables. 
\begin{eqnarray*}
D_{1}&=& nE_{w}[E_{X}[|f(X,w)|^{2} ]], \\
D_{2}&=& nE_{X}[|E_{w}[f(X,w)]|^{2}], \\
D_{3}&=&\sum_{i=1}^{n}E_{w}[|f(X_{i},w)|^{2} ],\\
D_{4}&=&\sum_{i=1}^{n}|E_{w}[f(X_{i},w)]|^{2},\\
D_{5}&=&E_{w}[\sqrt{n}\;\eta_{n}(w)].
\end{eqnarray*}
\end{defn}
Then, by using Lemma \ref{lemma:part}, it follows that 
\begin{eqnarray}
E[G]&=& S
+\frac{1}{2n}E[D_{2}],\label{eq:GGG}\\
E[T]&=& S-\frac{\beta\sigma^{2}}{n}E[D_{3}-D_{4}]
+\frac{1}{2n}E[D_{4}],\label{eq:TTT}\\
E[V]&=& E[D_{3}-D_{4}]\label{eq:VVV},\\
E[D_{5}]&=&\beta\sigma^{2}E[D_{3}-D_{4}]+(1/2)E[D_{1}-D_{3}].
\end{eqnarray}

We show that five expectation values $E[D_{j}]$ $(j=1,2,3,4,5)$ converge to 
constants. To show such convergences, 
it is sufficient to prove that each $D_{j}$
weakly converges to some random variable and that 
$E[(D_{j})^{1+\delta}]<C$ for some $\delta>0$ and constant $C>0$ \cite{Wvan}.

\begin{defn}
For a given constant $\epsilon>0$, 
a localized expectation operator $E_{w}^{\epsilon}[\;\;]$ is defined by 
\begin{equation}\label{eq:Ew-loc}
E_{w}^{\epsilon}[F(w)]=\frac{\displaystyle \int_{K(w)\leq\epsilon} F(w)\exp(-\beta H_{0}(w))\varphi(w)dw}
{\displaystyle \int_{K(w)\leq\epsilon} \exp(-\beta H_{0}(w))\varphi(w)dw}.
\end{equation}
Let $D_{i}^{\epsilon}$ $(i=1,2,3,4,5)$ be random variables that are defined by 
replacing $E_{w}[\;\;]$ by $E_{w}^{\epsilon}[\;\;]$.
\end{defn}

\begin{lemma} \label{lemma:out-zero}
Let $0<\delta<s/4-1$. 
For arbitrary $\epsilon>0$, $j=1,2,3,4,5$, 
\[
\lim_{n\rightarrow\infty}E[|D_{j}-D_{j}^{\epsilon}|^{1+\delta}]=0.
\]
\end{lemma}
\begin{proof}
We can prove five equations by the same way. Let us prove the case $j=3$. 
Let $L(w)=\sum_{i=1}^{n}|f(X_{i},w)|^{2}$. Because $f(x,w)$ is $L^{s}(q)$-valued
analytic function, $E[(\|L\|/n)^{1+\delta}]<\infty$. 
\begin{eqnarray*}
|D_{3}-D_{3}^{\epsilon}| & \leq & 
\frac{\displaystyle \int_{K(w)\geq\epsilon} L(w)
\exp(-\beta H_{0}(w))\varphi(w)dw}
{\displaystyle \int_{K(w)\leq\epsilon} 
\exp(-\beta H_{0}(w))\varphi(w)dw}\\
&\leq & 
\frac{\|L\|\;e^{-n\beta\epsilon+2\beta\sqrt{n}\|\eta_{n}\|}}
{\int_{K(w)\leq \epsilon}\exp(-\beta nK(w))\varphi(w)dw}\\
&\leq & C_{1}\;n^{d/2}\|L\|\exp(-n\beta\epsilon/2+(2\beta/\epsilon)\|\eta_{n}\|^{2})
\end{eqnarray*}
where we used 
$2\sqrt{n}\|\eta_{n}\|\leq (n\epsilon/2+ (2/\epsilon)\|\eta_{n}\|^{2})$
and $C_{1}>0$ is a constant.
From Lemma \ref{lemma:eta-bdd}, $E[\|\eta_{n}\|^{s/2}]\equiv C_{2}<\infty$, hence
by using $C_{3}=(8\epsilon^{2})^{s/4}C_{2}$,
\[
P(\|\eta_{n}\|^{2}\geq n/(8\epsilon^{2}))\leq C_{3}/n^{s/4}.
\]
Let $E[F]_{A}$ be the expectation value of $F(x)I_{A}(x)$ 
where $I_{A}(x)$ is the defining function of a set $A$, in other words,
$I_{A}(x)=1$ if $x\in A$ or $0$ if otherwise. 
\begin{eqnarray*}
E[|D_{3}-D_{3}^{\epsilon}|^{1+\delta}]&
= &E[|D_{3}-D_{3}^{\epsilon}|^{1+\delta}]_{\{\|\eta_{n}\|^{2}\geq n/(8\epsilon^{2})\}}\\
&&+ E[|D_{3}-D_{3}^{\epsilon}|^{1+\delta}]_{\{\|\eta_{n}\|^{2}<n/(8\epsilon^{2})\}}.
\end{eqnarray*}
The first term of the right hand side is not larger than 
$C_{3}E[\|L\|^{1+\delta}]/n^{s/4}$ and the second term is not larger than
$E[(C_{1}\|L\|)^{1+\delta}]n^{d/2}\exp(-n\beta\epsilon/4)$. 
Both of them converge to zero. 
\end{proof}

\section{Resolution of Singularities}\label{section:res-sing}

To study the expectation on the region $W_{\epsilon}$ we need 
resolution of singularities because $W_{0}$ contains singularities
in general. Let $\epsilon>0$ be a sufficiently small constant. Then 
by applying Hironaka's theorem \cite{Hironaka} to the real analytic function
$K(w)\prod_{j=1}^{k}\pi_{j}(w)\varphi_{1}(w)$, all functions $K(w)$, $\pi_{j}(w)$,
and $\varphi_{1}(w)$ are made normal crossing. In fact, 
there exist an open set $W_{\epsilon}^{*}\subset W^{*}$ 
which contains $W_{\epsilon}$, a manifold $U^{*}$,
and a proper analytic map $g:U^{*}\rightarrow W_{\epsilon}^{*}$ such that
in each local coordinate of $U^{*}$, 
\begin{eqnarray*}
K(g(u))&=& u^{2k}, \\
\varphi(g(u))|g(u)'|&=& \phi(u)|u^{h}|,
\end{eqnarray*}
where $k=(k_{1},...,k_{d})$ and $h=(h_{1},...,h_{d})$ are multi-indices 
($k_{j}$ and $h_{j}$ are nonnegative integers), 
$u^{2k}=\prod_{j}u_{j}^{2k_{j}}$, $u^{h}=\prod_{j}u_{j}^{h_{j}}$, 
$|g(u)'|$ is the absolute value of Jacobian determinant of $w=g(u)$, 
and $\phi(u)>0$ is a function of class $C^{\infty}$. 
Let $U=g^{-1}(W_{\epsilon})$. Since $g$ is a proper map and 
$W_{\epsilon}$ is compact, $U$ is also compact. 
Moreover, it is covered by a finite sum 
\[
U=\cup_{\alpha}U_{\alpha},
\]
where each $U_{\alpha}$ can be taken to be $[0,b]^{d}$ in each local coordinate
using some $b>0$, and
\[
\int_{W_{\epsilon}}
 F(w) \varphi(w)dw=\sum_{\alpha}\int_{U_{\alpha}} F(g(u))\phi_{\alpha}(u)|u^{h}|du,
\]
where $\phi_{\alpha}(u)\geq 0$ is a function of class $C^{\infty}$.
In this paper, we apply these facts to analyzing the singular regression problem. 
For resolution of singularities and its applications, see \cite{Hironaka} and
\cite{Atiyah},\cite{watanabe2001}. Lemma \ref{lemma:zeta} 
is directly proved by these facts \cite{Atiyah,Kashiwara,watanabe2001}. 
Moreover, the following lemma is simultaneously obtained. 

\begin{lemma}
The largest pole $(-\lambda)$ and its order $m$ of $\zeta(z)$ are given by 
\begin{eqnarray}
\lambda&=&\min_{\alpha}\min_{j}\Bigl(\frac{h_{j}+1}{2k_{j}}\Bigr),\label{eq:11}\\
m&=&\max_{\alpha}\#\Bigl\{j;\lambda=\frac{h_{j}+1}{2k_{j}}\Bigr\},\label{eq:22}
\end{eqnarray}
where, if $k_{j}=0$, $(h_{j+1}+1)/2k_{j}$ is defined to be $+\infty$ and
$\#$ shows the number
of elements of the set. Let $\{U_{\alpha^{*}}\}$ be the set of all local coordinates
that attain both $\min_{\alpha}$ in eq.(\ref{eq:11}) and $\max_{\alpha}$
in eq.(\ref{eq:22}). Such coordinates are referred to as the essential coordinates. 
\end{lemma}

For a given real analytic function $K(w)$, 
there are infinitely many different resolutions of singularities. 
However, $\lambda$ and $m$ 
do not depend on the pair $(U^{*},g)$. They are called birational invariants. 
By the definition of $K(w)$ in eq.(\ref{eq:K(w)}), 
there exists an $L^{s}(q)$-valued analytic function $a(x,u)$ on each local coordinate 
in $U^{*}$ such that
\[
f(x,u)=a(x,u)u^{k}
\]
and $E_{X}[|a(X,u)|^{2}]=2$. Therefore,
\[
H_{0}(g(u))=n\;u^{2k}-\sqrt{n}\;u^{k}\;\xi_{n}(u),
\]
where
\[
\xi_{n}(u)=\frac{1}{\sqrt{n}}\sum_{i=1}^{n}S_{i}\cdot a(X_{i},u)
+\frac{1}{\sqrt{n}}\sum_{i=1}^{n}u^{k}
\Bigl(1-\frac{a(X_{i},u)^{2}}{2}\Bigr).
\]
Then $E[\|\xi_{n}\|^{s/2}]<\infty$ and 
$E[\|\nabla\xi_{n}\|^{s/2}]<\infty$, because both $a(x,u)$ and $\nabla a(x,u)$
are $L^{s}(q)$-valued analytic function, where 
$\|\nabla \xi_{n}\|=\max_{j}\sup_{w}|\partial_{j}\xi_{n}(u)|$. 
The expectation operator $E_{u}[\;\;]$ on $U$ is defined so that it satisfies
$E_{w}^{\epsilon}[F(w)]=E_{u}[F(g(u))]$. Then 
\begin{eqnarray*}
D_{1}^{\epsilon}&=& nE_{u}[2u^{2k}], \\
D_{2}^{\epsilon}&=& nE_{X}[|E_{u}[a(X,u)u^{k}]|^{2}], \\
D_{3}^{\epsilon}&=&\sum_{i=1}^{n}E_{u}[|a(X_{i},u)|^{2} u^{2k}],\\
D_{4}^{\epsilon}&=&\sum_{i=1}^{n}|E_{u}[a(X_{i},u)u^{k}]|^{2},\\
D_{5}^{\epsilon}&=&E_{u}[\sqrt{n}\xi_{n}(u)u^{k}].
\end{eqnarray*}

\begin{lemma}\label{lemma:e-bdd} Let $s\geq 12$ and $0<\delta<s/6-1$. 
For $i=1,2,3,4,5$, there exists a constant $C>0$ such that 
$E[(D_{i}^{\epsilon})^{1+\delta}]<C$ holds. 
\end{lemma}
\begin{proof} Since $0\leq D_{4}^{\epsilon}\leq D_{3}^{\epsilon}$,
$0\leq D_{2}^{\epsilon}\leq D_{1}^{\epsilon}$, and
$|D_{5}^{\epsilon}|\leq (\|\xi_{n}\|^{2}+2D_{1}^{\epsilon})/2$, it is 
sufficient to prove $j=1,3$. 
The proof for $j=1,3$ can be done 
by the same way. Let us prove the case $j=3$. In $l=1,2,..,d$, 
at least one of $k_{l}\geq 1$. 
By using partial integration for $du_{l}$, 
we can show that there exists $c_{1}>0$ such that 
\begin{equation}\label{eq:u2k-bdd}
E_{u}[u^{2k}]\leq \frac{c_{1}}{n}\{1+\|\xi_{n}\|^{2}+\|\nabla\xi_{n}\|^{2}\}.
\end{equation}
Therefore by using $L=(1/n)\sum_{i=1}^{n}\|a(X_{i})\|^{2}$
and H\"{o}lder's inequality with $1/3+1/(3/2)=1$,
\begin{eqnarray*}
&&E[(D_{3}^{\epsilon})^{1+\delta}]
\leq   E[(c_{1}L(1+\|\xi_{n}\|^{2}+\|\nabla\xi_{n}\|^{2}))^{1+\delta}] \\
&& \leq  E[(c_{1}L)^{3+3\delta}]^{1/3}
E[(1+\|\xi_{n}\|^{2}+\|\nabla\xi_{n}\|^{2})^{(3+3\delta)/2}]^{3/2}.
\end{eqnarray*}
Since $E[\|a(X)\|^{s}]<\infty$, $E[\|\xi_{n}\|^{s/2}]<\infty$, 
and $E[\|\nabla\xi_{n}\|^{s/2}]<\infty$, 
this expectation is finite. 
\end{proof}

\section{Renormalized distribution}

\begin{defn} For a given function $h(u)$ on $U$, 
the renormalized expectation operator $E_{u,t}^{*}[\;\;|h]$ is defined by
\[
E_{u,t}^{*}[F(u,t)|h]
=
\frac{\displaystyle\sum_{\alpha^{*}}\int_{0}^{\infty}dt \int D(du) F(u,t)t^{\lambda-1}
e^{-\beta t+\beta \sqrt{t}\;h(u)}}
{\displaystyle \sum_{\alpha^{*}}\int_{0}^{\infty}dt \int D(du)t^{\lambda-1}
e^{-\beta t+\beta \sqrt{t}\;h(u)}},
\]
where $D(du)$ is a measure which is defined in eq.(\ref{eq:D(du)})
and $\sum_{\alpha^{*}}$ shows the sum of all essential coordinates.
Also we define
\begin{eqnarray*}
D_{1}^{*}(h)&=& E_{u,t}^{*}[2t|h], \\
D_{2}^{*}(h)&=& E_{X}[|E_{u,t}^{*}[a(X,u)\sqrt{t}]|^{2}|h],\\
D_{5}^{*}(h)&=& E_{u,t}^{*}[h(u)\sqrt{t}|h].
\end{eqnarray*}
\end{defn}

\begin{lemma}\label{lemma:d-star}
The following convergences in probability hold.
\begin{eqnarray*}
D_{1}^{\epsilon}-D_{1}^{*}(\xi_{n})&\rightarrow &0,\\
D_{2}^{\epsilon}-D_{2}^{*}(\xi_{n})&\rightarrow &0,\\
D_{3}^{\epsilon}-D_{1}^{*}(\xi_{n})&\rightarrow &0,\\
D_{4}^{\epsilon}-D_{2}^{*}(\xi_{n})&\rightarrow &0,\\
D_{5}^{\epsilon}-D_{5}^{*}(\xi_{n})&\rightarrow &0.
\end{eqnarray*}
\end{lemma}
\begin{proof}
These five convergences can be proved by the same way. We show 
$D_{3}^{\epsilon}-D_{1}^{*}(\xi_{n})\rightarrow 0$. 
Let $L(u)=(1/n)\sum_{i=1}^{n}|a(X_{i},u)|^{2}$. Since
$E_{X}[|a(X,u)|^{2}]=2$, 
\begin{eqnarray*}
|D_{3}^{\epsilon}-D_{1}^{*}(\xi_{n})|&\leq & |E_{u}[nL(u)u^{2k}]-E_{u}[E_{X}[a(X,u)^{2}]u^{2k}]|\\
&& +|E_{u}[2u^{2k}]
-E_{u,t}^{*}[2t|\xi_{n}]|.
\end{eqnarray*}
Let the first and second terms of the left hand side of this inequality
be $D_{6}$ and $D_{7}$ respectively.
Then 
\[
D_{6}\leq \|L-a(X)\|^{2} E_{u}[n u^{2k}].
\]
By the convergence in probability $\|L-a(X)\|\rightarrow 0$ and 
eq.(\ref{eq:u2k-bdd}), $D_{6}$ 
converges to zero in probability. From Lemma \ref{lemma:app1} and \ref{lemma:app2}
in Appendix, it is derived that 
\begin{equation}\label{eq:u2k-t}
|E_{u}[u^{2k}]-E_{u,t}^{*}[t|\xi_{n}]|\leq \frac{c_{1}}{\log n}
\frac{e^{2\beta\|\xi_{n}\|^{2}}}
{\min(\phi)^{2}}
\{1+\beta \|\nabla \xi_{n}\|\},
\end{equation}
which shows $D_{7}\rightarrow 0$ in probability. 
\end{proof}

\begin{lemma}\label{lemma:xi-xi}
For arbitrary function $h(u)$, the following equality holds. 
\[
D_{1}^{*}(h)=D_{5}^{*}(h)+\frac{2\lambda}{\beta}.
\]
\end{lemma}
\begin{proof}
Let $F_{p}(u)$ be a function defined by
\[
F_{p}(u)=\int_{0}^{\infty}t^{p}\;t^{\lambda-1}\;e^{-\beta t+\beta \sqrt{t}h(u)}dt.
\]
Then by using the partial integration of $dt$,
\[
F_{1}(u)=\frac{1}{2}h(u)F_{1/2}(u)+\frac{\lambda}{\beta}F_{0}(u).
\]
By the definition of $D_{1}^{*}(h)=E_{u,t}^{*}[2t|h]$ and
$D_{5}^{*}(h)= E_{u,t}^{*}[h(u)\sqrt{t}|h]$, we obtain the lemma. 
\end{proof}

\section{Proof of Main Theorems}\label{section:proof}

\subsection{Proof of Lemma \ref{lemma:zeta}}
\begin{proof}
Lemma \ref{lemma:zeta} is already proved in section \ref{section:res-sing}. 
\end{proof}

\subsection{Proof of Lemma \ref{lemma:sf}}
\begin{proof}
By the definition, $V=D_{3}-D_{4}$. 
By Lemma \ref{lemma:out-zero} and \ref{lemma:e-bdd}, 
$E[V^{1+\delta}]<\infty$. Reall that the convergence in law
$\xi_{n}\rightarrow \xi$ holds. 
The random variable
$D_{1}^{*}(\xi_{n})-D_{2}^{*}(\xi_{n})$ is a continuous function of $\xi_{n}$, 
hence it converges to a random variable $D_{1}^{*}(\xi)-D_{2}^{*}(\xi)$ 
in law. Therefore, 
by Lemma \ref{lemma:out-zero} and \ref{lemma:d-star},
$D_{3}-D_{4}$ converges to the same random variable in law. 
Hence $E[V]$ converges to a constant when $n$ tends to infinity. 
\end{proof}

\subsection{Proof of Theorem \ref{theorem:G-T}}

\begin{proof}
By the same way as proof of Lemma \ref{lemma:sf}, both
$E[D_{1}]$ and $E[D_{3}]$ converge to $E[D_{1}^{*}(\xi)]$ whereas both
$E[D_{2}]$ and $E[D_{4}]$ converge to $E[D_{2}^{*}(\xi)]$. 
From eqs.(\ref{eq:GGG}), (\ref{eq:TTT}), and (\ref{eq:VVV})
\begin{eqnarray*}
E[n(G-S)]&\rightarrow & \frac{1}{2}E[D_{2}^{*}(\xi)],\\
E[n(T-S)]&\rightarrow & -2\sigma^{2}\nu + \frac{1}{2}E[D_{2}^{*}(\xi)],\\
E[V]&\rightarrow & E[D_{1}^{*}(\xi)]-E[D_{2}^{*}(\xi)],
\end{eqnarray*}
where we used the definition of $\nu$, that is to say, 
$E[D_{1}^{*}(\xi)-D_{2}^{*}(\xi)]=2\nu/\beta$. From Lemma \ref{lemma:xi-xi},
\[
E[D_{1}^{*}(\xi)]=2\sigma^{2}\nu+\frac{2\lambda}{\beta},
\]
resulting that
\[
E[D_{2}^{*}(\xi)]=2\sigma^{2}\nu+\frac{2\lambda-2\nu}{\beta},
\]
which completes the theorem. 
\end{proof}

\subsection{Proof of Theorem \ref{theorem:eq-state}}

\begin{proof}
From Theorem \ref{theorem:G-T}, 
\begin{eqnarray*}
E[G]&=&\frac{N\sigma^{2}}{2}+
\Bigl(\frac{\lambda-\nu}{\beta}+\nu\sigma^{2}
\Bigr)\frac{1}{n}+o_{n}, \\
E[T]&=&\frac{N\sigma^{2}}{2}+
\Bigl(\frac{\lambda-\nu}{\beta}-\nu\sigma^{2}
\Bigr)\frac{1}{n}+o_{n},
\end{eqnarray*}
where $no_{n}\rightarrow 0$. Therefore
\begin{eqnarray*}
E[G]&=&E[T]+\frac{2\nu\sigma^{2}}{n}+o_{n}\\
&=&E[T]\Bigl(1+\frac{2\beta E[V]}{Nn}\Bigr)+o_{n}.
\end{eqnarray*}
To prove Theorem \ref{theorem:eq-state}, it is sufficient to show
$E[VT]-E[V]E[T]\rightarrow 0$. 
\[
E[|V(T-E[T])|]\leq E[V^{2}]^{1/2}E[(T-E[T])^{2}]^{1/2}.
\]
Since $s/4-1\geq 2$, 
\[
0\leq E[V^{2}]\leq E[(D_{3})^{2}]<\infty.
\]
Let $S^{(n)}=\frac{1}{n}\sum_{i=1}^{n}|S_{i}|^{2}/2$, $S=\sigma^{2}N/2$.
Then
\[
E[(T-E[T])^{2}]\leq 3E[(T-S^{(n)})^{2}+(S^{(n)}-S)^{2}+(S-E[T])^{2}].
\]
Firstly, 
from
\[
T-S^{(n)}=\frac{E_{w}[\eta_{n}(w)]}{\sqrt{n}}+\frac{D_{3}}{2n^{2}},
\]
we obtain 
\[
E[(T-S^{(n)})^{2}]\leq \frac{2E[\|\eta\|^{2}]}{n}+ \frac{E[D_{3}^{2}]}{n},
\]
which converges to zero. Secondly, 
$\{S_{i}\}$ are independently subject to the normal distribution,
hence 
$E[(S^{(n)}-S)^{2}]\rightarrow 0$. 
And lastly,
\[
T-S= \frac{D_{1}}{n},
\] 
hence $E[(T-S)^{2}]$ also converges to zero. 
\end{proof}

\section{Conclusion}

In this paper, we proved that singular regression problem is 
mathematically determined by two birational invariants, 
the real log canonical threshold and singular fluctuation.
Moreover, there is a universal relation between the generalization
error and the training error, by which we can estimate 
two birational invariants from random samples.

\section*{Appendix}

To prove eq.(\ref{eq:u2k-t}), we use the following lemmas. 
Let $\xi$ and $\varphi$ are functions of $C^{1}$ class from $[0,b]^{d}$ to
${\R}$. Assume that $\varphi(u)>0$, $u=(x,y)\in [0,b]^{d}$. The
partition function of $\xi$, $\varphi$, $n>1$, and $p\geq 0$ is defined by
\begin{eqnarray}
Z^{p}(n,\xi,\varphi)&=&\int_{[0,b]^{m}}dx
\int_{[0,b]^{d-m}}dy\nonumber \;
 K(x,y)^{p} \;x^{h}y^{h'}\;\varphi(x,y)\\
&& \label{eq:zp}
\times\exp(-n\beta \;K(x,y)^{2}+\sqrt{n}\beta \;K(x,y)\;\xi(x,y)).
\end{eqnarray}
where $K(x,y)=x^{k}y^{k'}$. Let us use 
\begin{eqnarray*}
\|\xi\|&=& \max_{(x,y)\in [0,b]^{d}}|\xi(x,y)|,\\
\|\nabla \xi\|&=&\max_{1\leq j\leq m}
\max_{(x,y)\in [0,b]^{d}}
\Bigl|
\frac{\partial \xi}{\partial x_{j}}
\Bigr|.
\end{eqnarray*}
Without loss of generality, 
we can assume that four multi-indices $k,k',h,h'$ satisfy
\[
\frac{h_{1}+1}{2k_{1}}=\cdots=
\frac{h_{r}+1}{2k_{m}}=\lambda<
\frac{h'_{j}+1}{2k'_{j}}
\;\;\;(j=m+1,2,...,d).
\]
In this appendix, we define 
$a(n,p)\equiv (\log n)^{m-1}/n^{\lambda+p}$.

\begin{lemma}\label{lemma:app1}
There exist constants $c_{1},c_{2}>0$ such that
for arbitrary $\xi$ and $\varphi$ ($\varphi(x)>0 \in [0,b]^{d}$)
and an arbitrary natural number $n>1$, 
\[
c_{1}\;a(n,p)
\;e^{-\beta \|\xi\|^{2}/2}
\min(\varphi)\leq
Z^{p}(n,\xi,\varphi)\leq 
c_{2}\;a(n,p)
\;e^{\beta \|\xi\|^{2}/2}
\;\|\varphi\|
\]
holds, where 
$\displaystyle
\min(\varphi)=\min_{u\in [0,b]^{d}}\varphi(u)$.
\end{lemma}

Let $\xi$ and $\varphi$ be functions of class $C^{1}$. We define
\[
Y^{p}(n,\xi,\varphi)\equiv \gamma\; a(n,p)
\int_{0}^{\infty}dt
\int_{[0,b]^{s}}dy\;
t^{\lambda+p-1}y^{\mu}
e^{-\beta t+\beta \sqrt{t}\xi_{0}(y)}\varphi_{0}(y),
\]
where we use notations, 
$\gamma=b^{|h|+m-2|k|\lambda}/(2^{m}(m-1)!\prod_{j=m+1}^{d} k_{j})$, 
$
\xi_{0}(y)=\xi(0,y)$, $\varphi_{0}(y)=\varphi(0,y)$, 
$\mu=h'-2\lambda k'$. A measure $D(du)$ on ${\R}^{d}$ is defined by
\begin{equation}\label{eq:D(du)}
D(du)=\gamma \delta(x)y^{\mu}.
\end{equation}

\begin{lemma}\label{lemma:app2}
There exists a constant $c_{3}>0$ such that, for arbitrary
$n>1$,  $\xi$,  $\varphi$, and $p\geq 0$, 
\begin{eqnarray*}
&&|Z^{p}(n,\xi,\varphi)-Y^{p}(n,\xi,\varphi)|\\
&&\leq \frac{c_{1}\;a(n,p)}{\log n}\;
e^{\beta\|\xi\|^{2}/2}\{\beta\|\nabla \xi\|\|\varphi\|+\|\nabla\varphi\|+\|\varphi\|\}.
\end{eqnarray*}
Moreover, there exist constant $c_{4},c_{5}>0$ such that, 
for arbitrary $\xi$, $\varphi$, $n>1$, 
\[
c_{4}\;a(n,p)
\;e^{-\beta\|\xi\|^{2}/2}
\min(\varphi)\leq
Y^{p}(n,\xi,\varphi)\leq 
c_{5}\;a(n,p)
\;e^{\beta\|\xi\|^{2}/2}
\;\|\varphi\|.
\]
\end{lemma}
\begin{proof}
Lemmas \ref{lemma:app1} and \ref{lemma:app2} are proved
by direct but rather complicated calculations \cite{watanabe2006,watanabe2009}. 
Let us introduce the outline of the proof. 
Let $F_{p}(x,y)$ be the integrated function in eq.(\ref{eq:zp})
and $Z^{p}=Z^{p}(n,\xi,\phi)$. 
\[
Z^{p}=\int dx\int dy F_{p}(x,y),
\]
which is equal to
\begin{equation}\label{eq:zzz-ppp}
Z^{p}=\int_{0}^{\infty}dt \int_{[0,b]^{d}} dx\;dy\; \delta(t-K(x,y)^{2})\;F_{p}(x,y).
\end{equation}
Therefore, the problem results in $\delta(t-K(x,y)^{2})$. 
For arbitrary function $\Psi(x,y)$ of class $C^{\infty}$, 
the function
\[
\zeta(z)=\int_{[0,b]^{d}} K(x,y)^{2z}\Psi(x,y)\;dxdy
\]
is the meromorphic function whose poles are 
$(-\lambda_{j})$ and its order $m_{j}$, hence it has
Laurent expansion, 
\[
\zeta(z)=\zeta_{0}(z)+\sum_{j=1}^{\infty}\frac{c_{j}(\Psi)}{(z+\lambda_{j})^{m_{j}}},
\]
where $\zeta_{0}(z)$ is a holomorphic function and $c_{j}(\Psi)$ is a 
Schwartz distribution. Since $\int\delta(t-K(x,y)^{2})\Psi(x,y)dxdy$ 
is the Mellin transform of $\zeta(z)$, we have an asymptotic expansion of 
$\delta(t-K(x,y)^{2})$ for $t\rightarrow +0$,
\[
\delta(t-K(x,y)^{2})=\sum_{j=1}^{\infty}\sum_{m=1}^{m_{j}}
t^{\lambda_{j}-1}(-\log t)^{m-1}c_{jm}(x,y),
\]
where $c_{jm}(x,y)$ is a Schwartz distribution. 
By applying this expansion to eq.(\ref{eq:zzz-ppp}), we obtain two lemmas. 
\end{proof}


\end{document}